\documentclass[letterpaper]{article} 
\usepackage{aaai23}  
\usepackage{times}  
\usepackage{helvet}  
\usepackage{courier}  
\usepackage[hyphens]{url}  
\usepackage{graphicx} 
\usepackage{amssymb}
\usepackage{amsmath}
\usepackage{booktabs}
\usepackage{array}
\usepackage{natbib}  
\usepackage{caption} 
\usepackage{algorithm}
\usepackage{algorithmic}
 
\def\ie{\emph{i.e.}}

\usepackage{newfloat}
\usepackage{listings}
\DeclareCaptionStyle{ruled}{labelfont=normalfont,labelsep=colon,strut=off} 
\floatstyle{ruled}
\newfloat{listing}{tb}{lst}{}
\floatname{listing}{Listing}
\title{Dual Memory Aggregation Network for Event-Based Object Detection with Learnable Representation}
\author{
    Dongsheng Wang\textsuperscript{\rm 1},
    Xu Jia\textsuperscript{\rm 1$\ast$},
    Yang Zhang\textsuperscript{\rm 1},
    Xinyu Zhang\textsuperscript{\rm 1},\\
    Yaoyuan Wang\textsuperscript{\rm 2},
    Ziyang Zhang\textsuperscript{\rm 2},
    Dong Wang\textsuperscript{\rm 1},
    Huchuan Lu\textsuperscript{\rm 1}\thanks{Corresponding authors.}
}
\affiliations{
    \textsuperscript{\rm 1}School of Artificial Intelligence, Dalian University of Technology, China\\
    \textsuperscript{\rm 2}Advanced Computing and Storage Lab, Huawei Technologies Co. Ltd, China\\
    \{32009141, cain, zhangxy71102\}@mail.dlut.edu.cn, 
    \{wangyaoyuan1, zhangziyang11\}@huawei.com, \\
    \{xjia, wdice, lhchuan\}@dlut.edu.cn

}

\usepackage{bibentry}

\begin{document}

\maketitle

\begin{abstract}
Event-based cameras are bio-inspired sensors that capture brightness change of every pixel in an asynchronous manner. Compared with frame-based sensors, event cameras have microsecond-level latency and high dynamic range, hence showing great potential for object detection under high-speed motion and poor illumination conditions.
Due to sparsity and asynchronism nature with event streams, most of existing approaches resort to hand-crafted methods to convert event data into 2D grid representation. However, they are sub-optimal in aggregating information from event stream for object detection. 
In this work, we propose to learn an event representation optimized for event-based object detection. Specifically, event streams are divided into grids in the $x$-$y$-$t$ coordinates for both positive and negative polarity, producing a set of pillars as 3D tensor representation.
To fully exploit information with event streams to detect objects, a dual-memory aggregation network (DMANet) is proposed to leverage both long and short memory along event streams to aggregate effective information for object detection. Long memory is encoded in the hidden state of adaptive convLSTMs while short memory is  modeled by computing spatial-temporal correlation between event pillars at neighboring time intervals. Extensive experiments on the recently released event-based automotive detection dataset demonstrate the effectiveness of the proposed method.
\end{abstract}

\begin{figure}[t]
    \centering
\includegraphics[width=8.0cm, height=5.0cm]{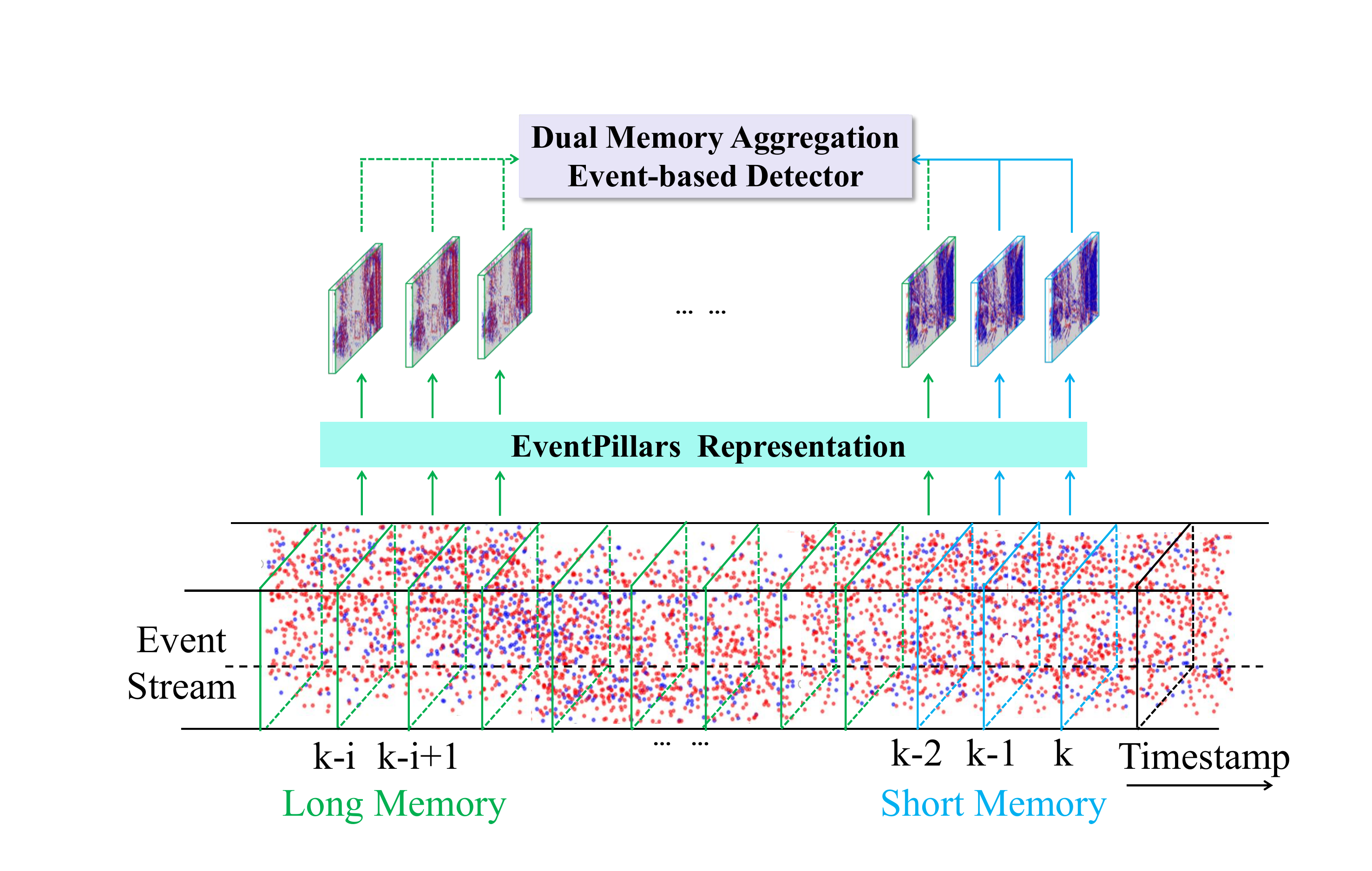}
    \caption{Illustration of the proposed method. At each time step, EventPillars is applied to convert event streams into a 3D grid representation as the input of DMANet. Then our network leverages both long and short spatial-temporal information with events to obtain the detection results of the $k$-th time step.}
    \label{fig:intro}
\end{figure}

\section{Introduction}

Object detection is regarded as the core basis of autonomous driving system. Generally, most of detection algorithms are designed based on frame-based sensors, which enable the system to better understand the real driving scenarios. These sensors are inexpensive and can provide rich information of scenes. However, due to the low frame rate, conventional cameras is not ideal for object detection under poor illumination conditions and fast motion. In addition, these cameras only capture static images, making it difficult for frame-based detectors to leverage motion information of scenes. 

With the advance of neuromorphic vision sensing (NVS) sensors, these limitations of frame-based cameras have been addressed. Event-based cameras, such as dynamic vision sensors (DVS) ~\cite{2014Retinomorphic}, are novel bio-inspired vision sensors that encode visual information in pixel-level event streams. In contrast to conventional frame-based cameras, which capture intensity images at a constant frame rate (e.g., 30fps), event cameras emit sequences of sparse and asynchronous events only at positions where brightness changes. Therefore, the event cameras have several advantages such as high dynamic range (HDR, exceeding 140 dB) ~\cite{2014Retinomorphic}, high temporal resolution, microsecond-level latency and are almost absent of motion blur. These properties make event cameras alone attractive in performing tasks such as object detection ~\cite{2019Event,perot2020learning}, tracking ~\cite{2018Asynchronous,2020Asynchronous,2018Event} and optical flow estimation ~\cite{2018EV,2018Unsupervised,2021Spatio}, even in challenging scenarios (e.g., low lighting conditions and high-speed moving scenes) ~\cite{2020Event}.

Output signal of an event camera is a stream of events, with each one recording an event's pixel location, time stamp and polarity, which are very different from conventional cameras. Therefore, it would be difficult to directly apply image-based object detection approaches to event streams. It requires novel representation and paradigm for object detection with event streams. Most of existing methods transform asynchronous event data to a synchronous image with several channels over a fixed time interval~\cite{2017Real,2017HOTS,2018Event,2018SegNet}. These methods encode the spatial and temporal information of events to facilitate its processing, and are easy to be applied to standard image-based deep learning algorithms. Voxel-based methods ~\cite{2019Unsupervised,perot2020learning,hu2020exploiting} discretizes event streams into different bins and groups them into a space-time histogram. However, these hand-crafted event representations discard sparsity of events and highly quantify timestamps of events, resulting in low performance among different vision tasks. Recent approaches Event Spike Tensor (EST)~\cite{2019End} and MatrixLSTM~\cite{2020Matrix} are directly aggregating information from raw event data through the learnable network. These methods have limitations in accuracy and are only for event-based object classification or optical flow estimation, thus they may not be applicable with object detection. 

In this paper, we propose two main contributions toward event-based object detection. First,
we present a novel event processing algorithm named EventPillars, which is end-to-end learnable representation for object detection. Compared to existing methods, EventPillars regards each event as a point in the 3D $x$-$y$-$t$ space and can be jointly trained to generate a compact 2D event representation for event streams over a time interval. 
In specific, EventPillars first divides event streams into a set of equally spaced pillars (along timestamps of events) and then generates a multi-channel event image by encoding each pillar with a 2D convolution network. In this way, EventPillars can take advantage of the spatio-temporal sparsity of events, outperforming existing state-of-the-art methods. 
Second, to make full use of streaming data to determine locations of objects, we propose an event-based object detection framework called Dual Memory Aggregation Network (DMANet), as shown in Fig.~\ref{fig:intro}. 
Different from image-based object detection methods, which discover objects in a static intensity frame, DMANet dynamically aggregates both long range and short range spatial-temporal information with the proposed Long Range Memory module (LRM) and Short Range Memory module (SRM) respectively. The proposed method is evaluated on an event-based automotive object detection dataset \cite{perot2020learning}. Extensive ablation study demonstrates the effectiveness of the proposed learnable event representation EventPillars and event-based object detection framework DMANet.
In summary, the contribution of this work are three-fold: 
\begin{itemize}
    \item EventPillars, a learnable event representation that converts asynchronous streams of events to synchronous event-based frames is proposed in this work. Compared to previous hand-crafted representations, EventPillars is able to aggregate more representative spatial-temporal information for object detection.
    \item A novel framework called DMANet, which  aggregates both short and long range history information over time, is proposed for for event-based object detection.
    \item The effectiveness of the proposed EventPillars and DMANet is validated in experiments on 1 Megapixel Automotive Detection Dataset. It outperforms state-of-the-art event-based object detectors by noticeable margin of 6.6\% mAP. 
\end{itemize}

\begin{figure*}[t]
\includegraphics[width=17cm,height=7cm]{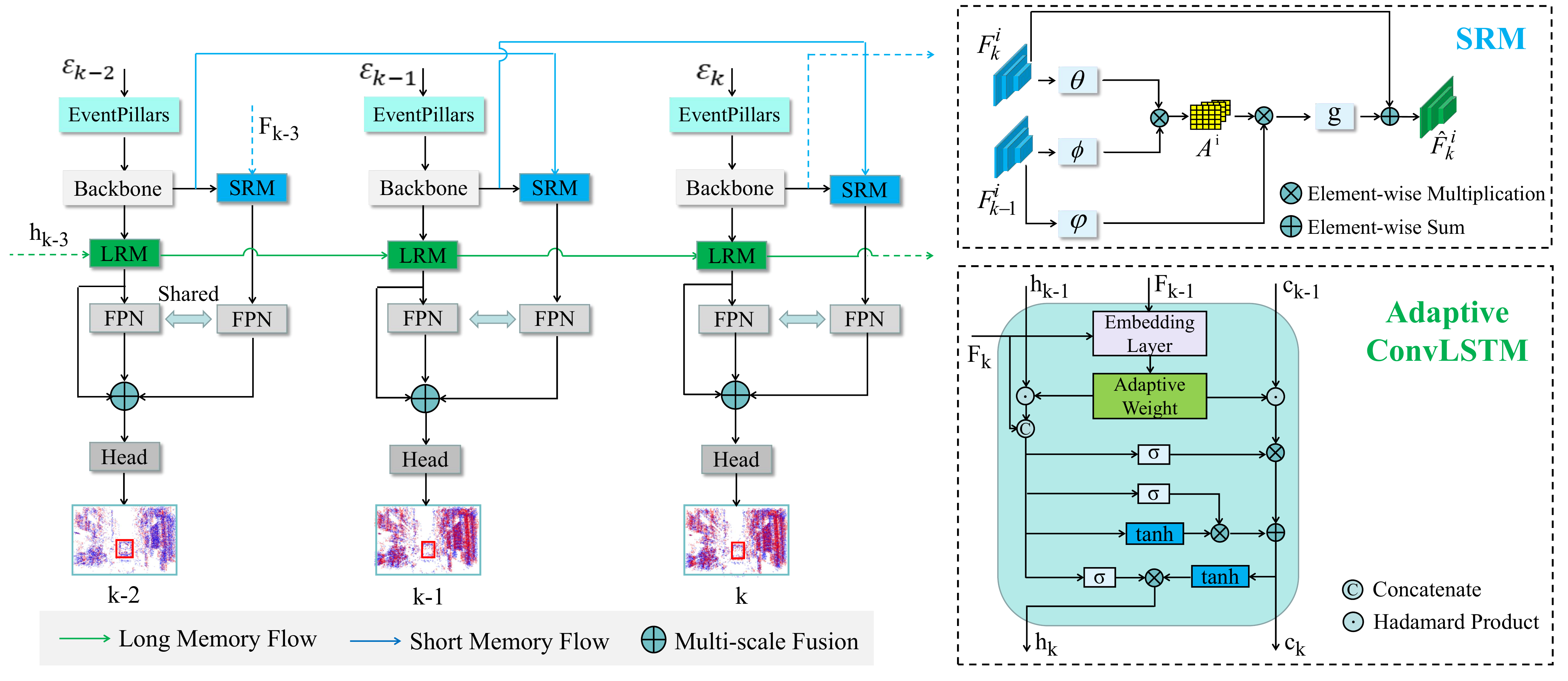}
  \centering
  \caption{The pipeline of the proposed architecture. Given a set of events $ \left \{ \varepsilon{\huge } _{i} \right \} _{i=0}^{k} $, EventPillars module converts events to a 3D tensor. Next, a CNN backbone is used to extract multi-scale features from the converted 3D event representations. 
  These features are further enhanced over time by aggregating spatial and temporal information within events with Long Range Memory Module (LRM) and Short Range Memory Module (SRM). Finally, the enhanced spatial-temporal features are fed into detection head to predict the bounding boxes.}
  \label{fig:overview}
\end{figure*}

\section{Related Work}
\noindent{\bfseries Event Representations.} 
Due to the asynchronous and sparse nature of events, event processing is of great importance in the event-based vision.
In recent years, several event-based representations have been proposed for various tasks. According to event aggregation strategies, these representations can be divided into hand-crafted transformations  ~\cite{2017Real,2018Event,2016Time,2017HOTS,2018SegNet,2018EV,2019Unsupervised}
and end-to-end learnable ones~\cite{2019End,2020Matrix}.
Hand-crafted feature representations are widely used in event-based vision community, which convert asynchronous event sequences to synchronous frames that can be directly fed into deep neural networks. 
However, most hand-crafted event representations heavily compress temporal information with events. 
Several learnable event representations have been proposed recently. 
EST ~\cite{2019End} learns a dense representation directly from raw event streams by applying an adaptive learning kernel instead of impulse functions. Matrix-LSTM ~\cite{2020Matrix} uses an LSTM network to process events and aggregate information into a 3D tensor. However, these event representations are mostly designed for task of object classification and may not be suitable for location-oriented tasks such as object detection. 
In this work, we propose EventPillars to preserve sparsity of events while capturing typical spatial-temporal information with events for the task of object detection in a learnable way.

\noindent{\bfseries Event-based Object Detection.} 
Most relevant works are event-based object detection approaches which are developed very recently. Hu et al. ~\cite {hu2020exploiting} proposed Network Grafting Algorithm (NGA) by utilizing powerful deep models pre-trained on images as initialization, which can save training costs and greatly reuse the existing frame-based datasets. 
The fcYOLO~\cite{cannici2019asynchronous}
converts events into leaky surface and introduces a hybrid network that can extract features in a sparse and asynchronous way. 
Messikommer et al. ~\cite{messikommer2020event} designed a novel event-based processing framework to transform a synchronous network into an asynchronous one which can process events asynchronously and greatly reduce computation. Perot et al. ~\cite{perot2020learning} introduced a recurrent event-based object detector with a temporal consistency loss for better regression.
Compared to RED~\cite{perot2020learning}, we apply LRM and SRM to aggregate both short and long range spatial-temporal information which is favorable for object detection.

\section{Methods}
\subsection{Overview}
Given an event camera with the spatial resolution of $ H\times W $, it outputs a stream of events whenever the log intensity changes over a set threshold $ C $. The events are denoted as $ \{e_{i}\}^{\tau} = \left ( x_{i} , y_{i}, t_{i}, p_{i} \right )$, where $ (x_i, y_i)\in \left [ 0, H \right ] ~\times \left [ 0, W \right ] $ are spatial coordinate of a pixel, $ t_{i} $ is the timestamp of the event, $ p_{i} \in \left \{ -1, 1 \right \} $ is the polarity of the change in brightness and $ \tau $ is a time interval.
Given a stream of events $ \{e_{i}\}^{\tau}$, the task of object detection is to predict precise bounding boxes 
for objects in the scene over the short interval and classify them into correct object categories.
Since the event camera has very high temporal resolution, quantities of events could be generated in a short interval, making it expensive and time-consuming to process every incoming event one by one. In addition, to make it applicable to powerful convolutional neural networks, event streams should be converted to grid-like representation like frames. 
In this work, we propose a novel learnable event representation called EventPillars, as shown in Fig.~\ref{fig:eventpillars}. In addition, a Dual Memory Aggregation Network (DMANet) is proposed for event-based object detection
by aggregating both long and short range spatial-temporal information. The overall framework for event-based object detection is shown in Fig.~\ref{fig:overview}. 

\subsection{EventPillars}
\label{sec:EventPillars}
Due to the sparsity and asynchronicity of event streams, computing suitable representation for events is the first step in the event-based vision algorithms. Most of the previous approaches apply hand-crafted event representation to encode spatial-temporal information with event data and then convert events into a multi-channel image. However, we notice that these representations are highly sensitive to speed-variant motion scenes, especially for movement-to-still moments in driving scenes. That is because most hand-crafted representations are spatially sparse, which makes them less robust to the change of number of events.
Although some learning-based event representations~\cite{2019End,2020Matrix} have been proposed recently, they are only validated on object classification and may not be suitable for location-oriented tasks like object detection. 

\begin{figure}[t]
\includegraphics[width=8cm, height=3.5cm]{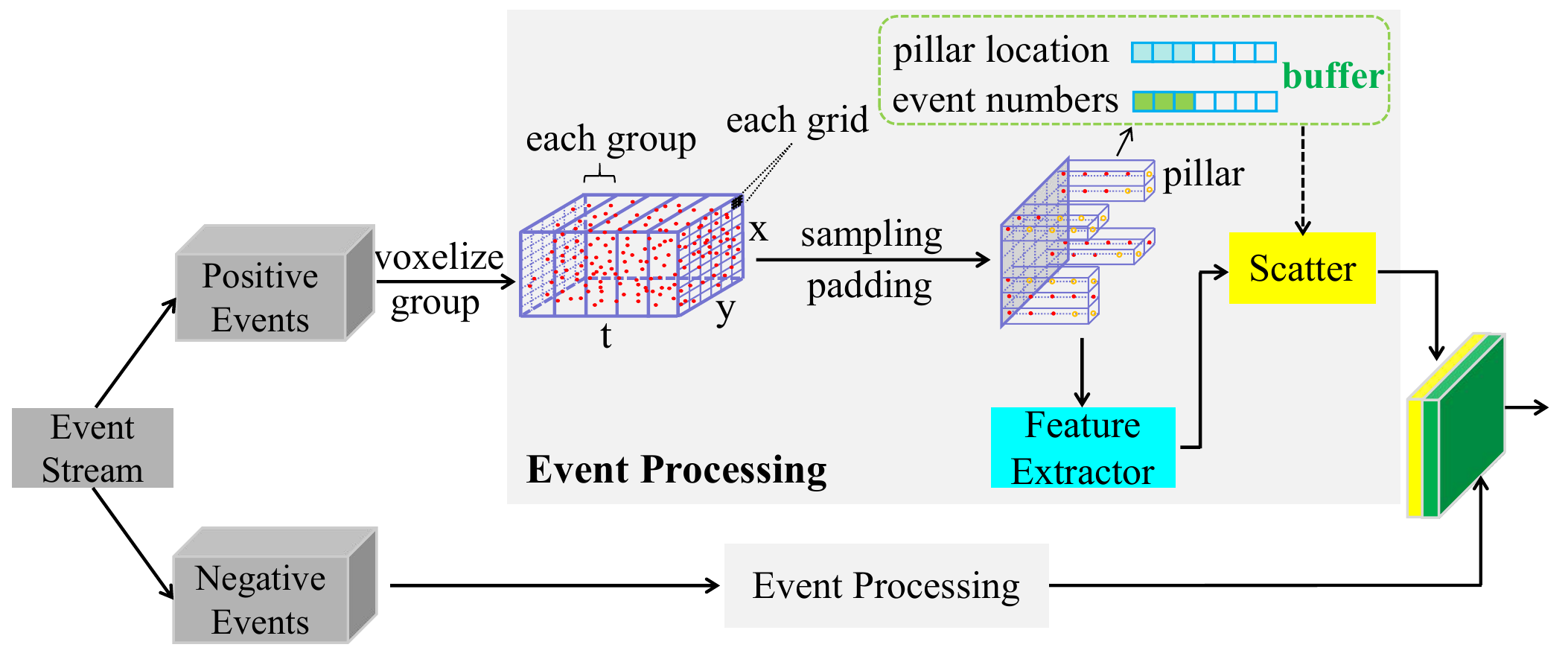}
    \centering
    \caption{The pipeline of EventPillars Representation.}
    \label{fig:eventpillars}
\end{figure}

Motivated by the success of learnable representation PointPillars on point cloud~\cite{2019PointPillars}, we also propose to compute event representation with a learnable encoder instead of relying on a fixed one. The design of EventPillars is shown in Fig.~\ref{fig:eventpillars}. Event representations for positive events and negative events are computed separately but in a similar way. For positive events, we first divide them into a series of equally spaced grids in the $x$-$y$ plane and each grid has a size of $1\times1$ in our experiments.
Since directly processing millions of events would dramatically increase computation cost, we group events into $D$ slices along time and then randomly sample a few events for each grid in a group.
We then set a fixed number of pillars ($K$) and a fixed number of events ($M$) per pillar. 
Spatial coordinates of pillars and number of events in it are recorded. 
Note that dense representation ($ \mathbb{R}^{ K \times M\times C} $), coordinates of pillars ($\mathbb{R}^{ K \times 3} $) and event numbers per pillar ($\mathbb{R}^{ K \times 1} $) are initialized with zero.
For each pillar, it is a container that receives events within the corresponding spatial grid in chronological order. Events in that container would be randomly discarded if the number of events in the pillar is more than $M$.
For all remained events in those pillars, we augment each of them $(x, y, t) $ with three additional dimensions $ (x-x_c, y-y_c, t-t_c)$, where subscript $c$ denotes the mean of those remained events. 
In this way, we are able to create a dense tensor representation $ T_{1} \in \mathbb{R}^{ K \times M\times C} $, where $C$ is 6, $K$ and $M$ are set manually. 

Next, we apply a 1D convolution operation followed by a batch normalization and a relu operation to generate $ T_{2} \in \mathbb{R}^{ K \times M\times C^{'}} $. Then a max-pooling operation is taken to generate $ T_{3} \in \mathbb{R}^{K \times C^{'}} $.
Now we have obtained pillar features and coordinates of each pillar, we can map the features of pillars to a pseudo-image $
T_{o} \in \mathbb{R}^{ H\times W \times C^{'}} $ based on the location of pillars, which called Scatter operation.
The implementation of the EventPillars’s output formulated as:
\begin{align}
  T_{2} &= Conv_{1\times1}\left ( BN\left ( ReLU(T_{1}) \right )  \right ), T_{2} \in \mathbb{R}^{ K \times M \times C^{'}}\\
  T_{3} &= MaxPool\left ( T_{2} \right ), T_{3} \in \mathbb{R}^{K \times C ^{'}}\\
  T_{o} &= Scatter\left ( T_{3} \right ), T_{o} \in \mathbb{R}^{ H \times W \times C^{'}}
\end{align}
In this way, our EventPillars preserves more spatial-temporal information of events than hand-crafted representations, providing a more representative feature for event-based object detection. 

\begin{table*}[ht]
  \centering
  \small {
  \begin{tabular}{l c cc}
    \toprule
    Event Representation &Description & mAP@0.5 & mAP \\
    \midrule
    Event Frame~\cite{2017Real} & Image of event polarities& 37.1 &17.7\\
    Event Count~\cite{2018Event} & Separate images of events for both polarities & 37.4 &17.7\\
    Timestamp Image~\cite{2016Time} & Image of  most recent timestamp & 38.9 &18.9\\
    Time Surface~\cite{2017HOTS} & Exponential of newest timestamps&  35.6 &17.0\\
    EV-SegNet~\cite{2018SegNet} & Image of filtered timestamps \& event count & 40.0 &19.4\\
    Voxel Grid~\cite{2019Unsupervised} & Discretized event volume via binning & 39.7 &18.8 \\
    MatrixLSTM~\cite{2020Matrix} & Learned with LSTM  & 36.5  & 18.0\\
    \textbf{EventPillars (Ours)} & Learned discretized pillars via sampling &  \textbf{42.1} &\textbf{21.3}\\ 
  \bottomrule
\end{tabular}}
  \caption{Object detection performance (mAP@0.5 and mAP@0.5:0.95) of different event representations on the 1 Mpx Auto-Detection Sub Dataset. B and C denotes the number of temporal bins and channels, respectively.}
  \label{tab:rep}
\end{table*}

\subsection{Dual Memory Aggregation Network}
\label{sec:Dual Memory Aggregation Network}
Although event cameras show certain robustness under challenging scenarios, 
there are few event-based object detection algorithms. 
Here we propose a novel object detection framework for event streams, \ie, Dual Memory Aggregation Network (DMANet). It consists of a Long Range Memory (LRM) module and a Short Range Memory (SRM) module. Both memory modules work together to extract and aggregate informative spatial-temporal features for object detection.
Input of this network is a sequence of tensor representations $ \left \{T_{k} \right \} \in \mathbb{R}^{ H \times W \times 2C'} $ computed using EventPillars.

\noindent{\bfseries Short Range Memory Module.} Events are almost continuous in time and  moving trajectories of objects are recorded. To model spatial-temporal correlation between consecutive time intervals, we propose a Short Range Memory (SRM) module. It combines informative features of both previous and current time intervals such that temporal continuity with object movement could be fully exploited. 

The design of the SRM module is shown in Fig.~\ref{fig:overview}. 
Specifically, at the $ k $-th time interval,
a tensor representation $ T_{k} $ produced by EventPillars is taken as input of the backbone with several ResBlocks~\cite{2016Deep} to compute multi-scale features $
\left \{F_{k}^i \right \}_{i=1,2,3}\in \mathbb{R}^{ h_{i} \times w_{i}\times c_{i}} $. 
To make full use of spatial-temporal correlation between $
F_{k}^i $ and short-range memory $ F_{k-1}^i $,
for each spatial scale $i$ we first apply $1 \times 1$ convolutions to transform features of two time intervals, 
and then compute a temporal attention map $ A_{i} \in \mathbb{R}^{h_{i}w_{i} \times h_{i}w_{i}} $ through a softmax layer similar to~\cite{2017Non} as follows: 
  \begin{equation}
      A_{i} = softmax\left ( \frac{\theta\left (F_{k}^{i}\right )\ast \phi \left ( F_{k-1}^{i} \right )^{T}}{\sqrt{\frac{c_{i}}{r} } }\  \right ) ,
  \end{equation} where \begin{math}
    \frac{c_{i}}{r} 
  \end{math} is the reduced number of channels and \begin{math} T
  \end{math} denotes a transpose operation.  
Such an attention map is able to model spatial-temporal correlation between consecutive temporal intervals.
   Finally, the output of SRM module is  temporally aggregated features $ \hat{F_{k}^{i}} \in \mathbb{R}^{h_{i} \times w_{i}\times c_{i}} $ which is computed by applying temporal attention to features from previous interval $F_{k-1}^{i}$  as follows:
  \begin{equation}
      \hat{F_{k}^{i}} = F_{k}^{i} + A_{i}\ast \varphi \left ( F_{k-1}^{i} \right ) 
  \end{equation} Therefore, in the SRM module, information from the previous interval 
  $k-1$
  can be instantly integrated into the events in the following interval 
$k$.
 
\noindent{\bfseries Long Range Memory Module.} Since events are triggered once brightness change exceeds a threshold $ \delta $, there would be few events when there is no motion in the scene.
In this case, only focusing on information from recent memory could not promote integration of  spatial and temporal information.
Hence, we further design a module named Long Range Memory (LRM) to capture informative features aggregated over longer range of time for object localization.

The core module of the proposed LRM is a novel variant of ConvLSTM~\cite{2015Convolutional}, which is different from recurrent achitecture proposed by ~\cite{perot2020learning}. A ConvLSTM layer can preserve history information through its hidden state and cell unit, hence being widely used in different event-based vision tasks \cite{2019High,cadena2021spade,2020Matrix}. 
However, as for the task of event-based object detection in autonomous driving scenario, it is not trivial to directly fuse features of both past and present for localizing objects.
Object movement would cause misalignment between features in the past and current time step, which makes it difficult to effectively aggregate discriminative features about objects and ignore irrelevant features in the background.
To address this issue, we propose Adaptive-ConvLSTM, which can adaptively capture long temporal dependencies and aggregate spatial-temporal features over long range in time, as illustrated Fig.~\ref{fig:overview}. At each time step $k$, an embedding layer consisting of $ 3 \times 3
$ convolutions is applied to the features $ F_{k-1}$ and $ F_{k} $. An adaptive weight is computed to measure the importance of previous memory units $ h_{k-1}$ and $ c_{k-1}$ through cosine similarity. If the features $ F_{k-1}^i$ and $ F_{k}^i $ are highly relevant, a larger weight would be assigned to the previous memory units. In this way, the Adaptive-ConvSTM module not only can adaptively distill the informative features contained in past time steps, but also build temporal dependencies in the long temporal range. This would benefit the detection process even in case of little events are triggered.

\begin{table}
  \centering
  \small{
  \begin{tabular}{p{2.6cm} c c}
    \toprule
    Method & mAP@0.5 \// mAP & params \// runtime \\
    \midrule
    RetinaNet-18 
    
    ~\cite{2017Focal} & 26.4 \// 13.9  &20.1M \// 25.6ms \\
    RetinaNet-34 
    
    ~\cite{2017Focal} & 34.1 \// 18.1  &30.1M \// 32.4ms\\
    E2Vid-RetinaNet 
    
    ~\cite{2019High} & 37.3 \// 20.0 &30.8M \// 111.5ms \\ 
    RED\_est 
    
    ~\cite{2019End} & 35.7 \// 17.6 & 27.9M \// 44.0ms \\
    RED 
    
    ~\cite{perot2020learning} &39.7 \// 18.8 &27.9M \// 39.8ms \\
    \textbf{DMANet\_voxel (Ours)} &\textbf{44.4} \// \textbf{22.7}  &\textbf{28.2M} \// \textbf{29.4ms}\\
    \textbf {DMANet (Ours)} & \textbf{46.3} \// \textbf{24.7}  &\textbf{28.2M} \// \textbf{30.2ms} \\
  \bottomrule
\end{tabular}}
  \caption{Evaluation on the 1 Mpx Auto-Detection Sub Dataset.}
  \label{tab:det}
\end{table}

\section{Experiments}
In this section, we first introduce the experimental setup for the event-based object detection task. Then we present the implementation details of the proposed EventPillars and Dual Memory Aggregation Network (DMANet). In addition, we also perform an extensive comparative evaluation of event representations for event-based object detection and compare the proposed DMANet with other state-of-the-art event-based vision methods. 
We also provide both visualization of detection results as well as visualization.
Finally, we conduct some ablation studies on a publicly available event-based Automotive Detection Dataset ~\cite{perot2020learning}.

\begin{figure*}[htp]
    \centering
\includegraphics[width=16cm,height=7.9 cm]{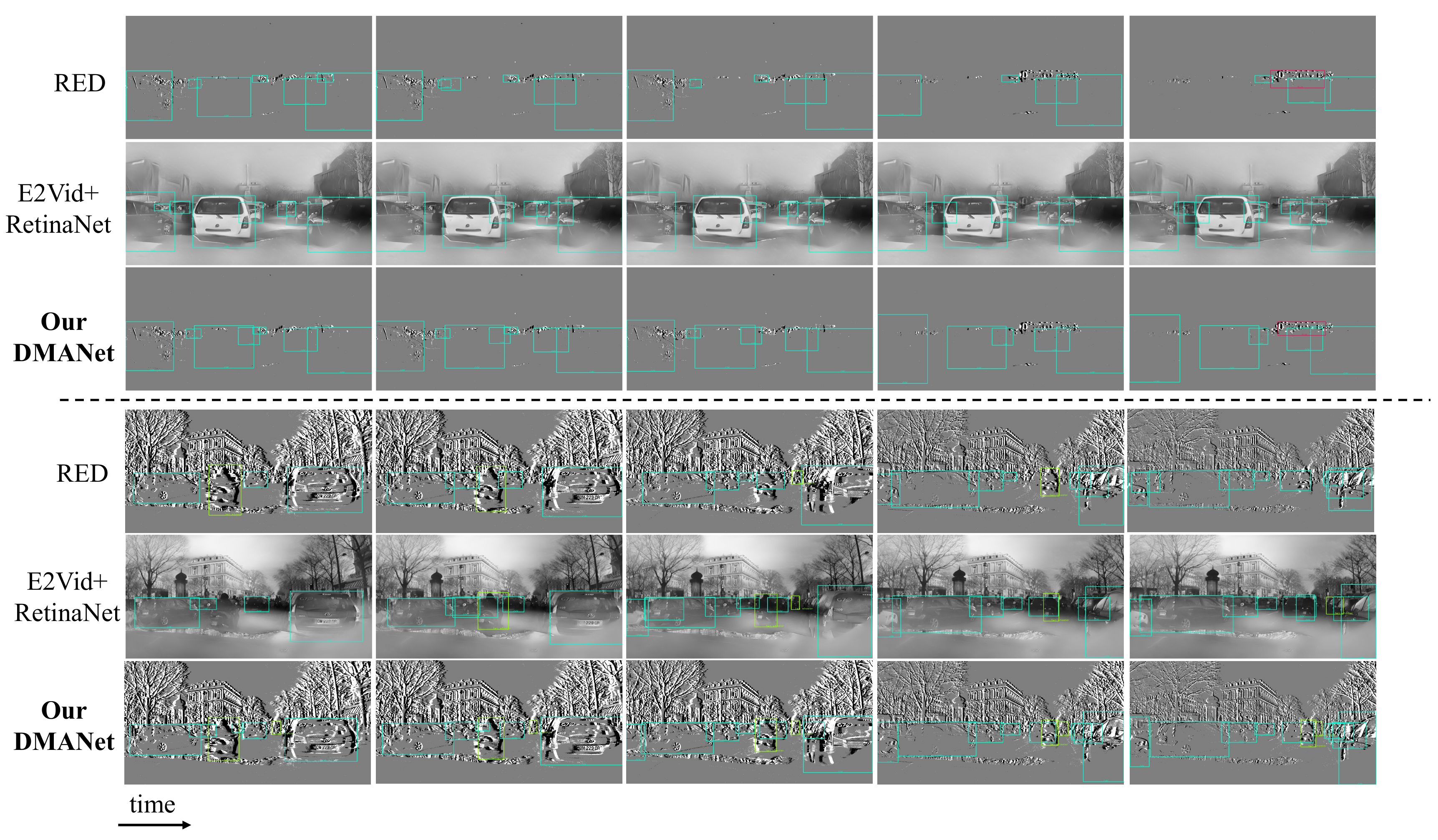}
    \caption{The qualitative results of RED, E2Vid-RetinaNet and DMANet in different scenarios.}
    \label{fig:visualize}
\end{figure*}

\subsection{Experimental Setup}  
\label{sec:Experimental Setup}
{\bfseries 1 Mpx Auto-Detection Sub Dataset.}  
Although event cameras have great potential in object detection under poor illumination conditions as well as high-speed motion, there are only a few available datasets for object detection. 
Prophesee was the first to release a high-resolution large-scale dataset for event-based object detection, with a $1280~\times720 $ event camera~\cite{EventSensor}. 
However, after visualizing event data in the dataset, we found that there are several problems with the dataset, such as $mosaicing$ events and incorrect labels, which may cause difficulty in model training. Therefore, we conduct data cleaning and sample a subset of challenging scenes from the original 
1 Megapixel Automotive Detection Dataset. We named this new dataset \textbf{1 Mpx Auto-Detection Sub Dataset}.
It has over 250 GB high quality event data after compression. The ratio of training set, validation set and test set is 0.70 : 0.15 : 0.15, which is similar to dataset used by the original authors.
All the experiments are conducted on this dataset in the following.

\noindent{\bfseries Evaluation Metrics.} We follow the previous work ~\cite{perot2020learning} to evaluate the performance of the proposed methods using the COCO metrics~\cite{lin2014microsoft}, and we report mAP@0.5 and mAP@0.5:0.95 respectively. We then select the best model on the validation set and apply it to the test set to report the final mAP. 

\subsection{Implementation Details}
\label{sec:Implementation Details}
{\bfseries Implementation of EventPillars.} In our experiments, we set the time interval of event streams to 50ms, which is same as ~\cite{perot2020learning}.
For each pillar, we set the maximum number of events per pillar $M$ to 5 and the number of non-empty pillars $ K $ to 100k. In addition, the output channels of EventPillars are set to 16. In order to make a fair comparison for other event processing strategies, the output of all event representations are uniformly resized to 512 $\times$ 512 in both training and inference phase. Note that no extra data augmentation is applied. 

\noindent{\bfseries Implementation of DMANet.} The proposed DMANet is built based on Pytorch. We use ResNet-18~\cite{2016Deep} as the backbone of our network. As for the optimization, we use Adam optimizer with an initial learning rate of 2e-4 and a cosine annealing scheme for learning rate scheduling~\cite{loshchilov2016sgdr}. The time step for training adaptive ConvLSTMs is empirically set to 10 according to GPU memory.
All the experiments are conducted on a single NVIDIA RTX 2080Ti GPU. 
The code is available at \url{https://github.com/wds320/AAAI_Event_based_detection}.

\subsection{Effectiveness of the Proposed Method}
\label{sec:Effectiveness of the Proposed Method}
{\bfseries Effectiveness of EventPillars.} We evaluate the proposed event representation method EventPillars on the 1 Mpx Auto-Detection Sub Dataset and compare it with several hand-crafted event representations, as shown in Tab.~\ref{tab:rep}. 
Since there is no publicly available code for event-based object detection approaches, we reproduce the RED method~\cite{perot2020learning} by ourselves with help from the authors. It is taken as the baseline to evaluate the performance of different event representation methods.

As shown in Tab.~\ref{tab:rep}, the proposed EventPillars representation achieves the best performance among the compared event representation methods. That could be attributed to its capability in effectively aggregating spatial-temporal information from event streams. In addition, end-to-end learning fashion provides much flexibility and allows the EventPillars representation to fully adapt to the task of event-based object detection. In particular, the proposed EventPillars outperforms the widely adopted Voxel Grid method by 2.4\% w.r.t mAP@0.5. Simple 2D grid representations based on histogram of events like Event Frame and Event Count give very poor results due to their heavy compression of event in temporal dimension. Loss of temporal information make them not able to sufficiently leverage temporal continuousness with object movement. 
Time Surface performs worst with only 35.6\% mAP because it is very sensitive to the choice of exponential decay kernel function and noisy events. 
MatrixLSTM is also an end-to-end representation, learning a dense representation from raw events directly, but only achieves 36.5\%mAP due to limit in preserving spatial-temporal information within events.
Generally, event representations with multiple channels such as Voxel Grid, EV-SegNet and EventPillars show better performance than 2D grid image-like representations such as Event Frame, Event Count and Time Surface.

\noindent{\bfseries Effectiveness of DMANet.} 
Tab.~\ref{tab:det} shows the quantitative comparison results in terms of two evaluation metrics on the 1 Mpx Auto-Detection Sub Dataset. Although the 1 Mpx Dataset was recorded together with a RGB camera, only event data are released by the author, thus we do not compare our event-based detector with other image-based algorithms.
Note that here all detectors use the same Voxel Grid event representation as used in~\cite{2019Unsupervised} for fair comparison except for the proposed DMANet which uses EventPillars event representation and the method EST. Since ResNet-18 and ResNet-34 are not equipped with temporal information modeling capability and only spatial information within events are exploited, they perform worst among the compared methods. By directly aggregating stream of events into reconstructed images with E2Vid~\cite{2019High}, E2Vid-RetinaNet is able to implicitly leverage both spatial and temporal information within events and achieves better performance. By explicitly exploiting temporal information within events in the object detector, the RED model~\cite{perot2020learning} and the proposed methods are able to make further improvement on detection performance. 
DMANet, the proposed method with the same event representation as RED, outperforms state-of-the-art method RED by 4.7\% w.r.t. mAP with only 0.3M increase on the number of parameters. That could be attributed to the effectiveness of dual memory mechanism, i.e., LRM and SRM, in extracting informative features for the task of object detection. By replacing Voxel Grid representation with EventPillars, there is another gain of 1.9\% mAP in object detection performance. 

\subsection{Visualization of Detection Results} 
\label{sec:Visual Detection Results}
As shown in Fig.~\ref{fig:visualize}, we show the visual detection results of previous methods RED ~\cite{perot2020learning}, E2Vid~\cite{2019High} with RetinaNet and the proposed DMANet sequentially by row. In the first case (corresponding to the first to third rows), there is no relative motion between event camera and scene, thus only few events are generated. It is obvious that RED performs well at the first time step, but the performance drops dramatically as time went by. Compared with RED, our model can effectively aggregate both long range and short range spatial-temporal information from events. 
Thus, the proposed DMANet can continue detecting objects even when stopping generating events for a long period. E2Vid-RetinaNet first reconstruct grayscale image from events through the memory unit of ConvLSTM, and then uses RetinaNet to detect objects. This method highly depends on the quality of reconstruction and not trivial to be trained in an end-to-end manner.
In the second case (corresponding to the top three rows), some two-wheelers are moving fast in front of the car. It can be seen that neither E2Vid-RetinaNet nor RED can detect two-wheelers as well as heavy-occluded vehicles. However, the proposed DMANet can still detect these objects correctly.

\begin{table}
  \centering
  \small{
  \begin{tabular}{ccccc}
    \toprule
    Baseline & LRM & SRM & Skip-sum & mAP@0.5  \\
    \midrule
    \checkmark&  & & & 39.9\\
    \checkmark &\checkmark & & &42.0\\
    \checkmark & \checkmark &  &\checkmark &43.4\\
     &  & \checkmark & &41.3\\
    \checkmark &\checkmark &\checkmark & \checkmark&44.4\\
  \bottomrule
\end{tabular}}
  \caption{Ablation study on effectiveness of each component in our DMANet.}
  \label{tab:abl_net}
\end{table}

\subsection{Ablation Study}
\label{sec:Ablation Study}
\noindent{\bfseries Effectiveness of each component.} To demonstrate the effect of key components in the proposed method, we conduct a series of ablation experiments on the 1 Mpx Auto-Detection Sub Dataset. In ablation experiments, ResNet18 is used as the backbone network, and we add three ConvLSTM layers before FPN layer as our base detector.
Long Range Memory (LRM) and Short Range Memory (SRM) are two key components of the proposed method.
As illustrated in Tab.~\ref{tab:abl_net}, the baseline model achieves 39.9\% mAP on 1 Mpx Auto-Detection Sub Dataset. By only using LRM module instead of ConvLSTM, we boost mAP with an additional +2.1\%, demonstrating that LRM can effectively accumulate relevant information during a time interval. Next, by using the single frame detector RetinaNet-18 along with SRM module, we achieve 41.3 \%mAP, which shows that the strength of capturing spatial-temporal correlation features of SRM.
And then, we add the skip-sum layers to fuse low-level and high-level temporal features, the mAP has reached 43.4\%. 
Finally, by adding SRM module, we boost the mAP with an additional +1.0\%, achieving 44.4\%. These improvements show the effects of individual components of the proposed approach.

\begin{table}
  \centering
  \small{
  \begin{tabular}{c|ccc}
    \toprule
    Method & mAP@0.5 & mAP & params  \\
    \midrule
     a  & 40.7    & 20.9         & 47.5M \\
     b  & 41.0    & 21.0         & 47.5M \\
     c  & 42.0    & 21.1         & 26.8M \\
  \bottomrule
\end{tabular}}
  \caption{The performance of different combinations of method in LRM.}
  \label{tab:lrm}
\end{table}

\noindent{\bfseries Effectiveness on feature integration strategy in LRM.} In order to explore the effectiveness of different combination methods of LRM, we conduct three experiments and report mAP as well as parameters of models, as shown in Tab.~\ref{tab:lrm}. We use the RetinaNet with ResNet-18 as the base model for fair comparison. First, we combine several LRM modules after multi-scale outputs of FPN in \textbf{method a}, each LRM module is connected with a $3\times3 $ convolution with stride of 2. 
As for \textbf{method b}, we only use the largest scale feature of FPN instead of multi-scale features. In \textbf{method c}, we place LRM modules before FPN and the input of each LRM is different scale of features from Resblock. From Tab.~\ref{tab:lrm}, we  observe that different strategies of combination LRM can obtain good mAP. However, since parameters of ConvLSTM increases exponentially with input channels, \textbf{method a} and \textbf{b} have larger parameters than \textbf{method c}. Thus, we choose \textbf{method c} as in final implementation. 

\begin{figure}[t]
\includegraphics[width=8.0cm, height=4.0cm]{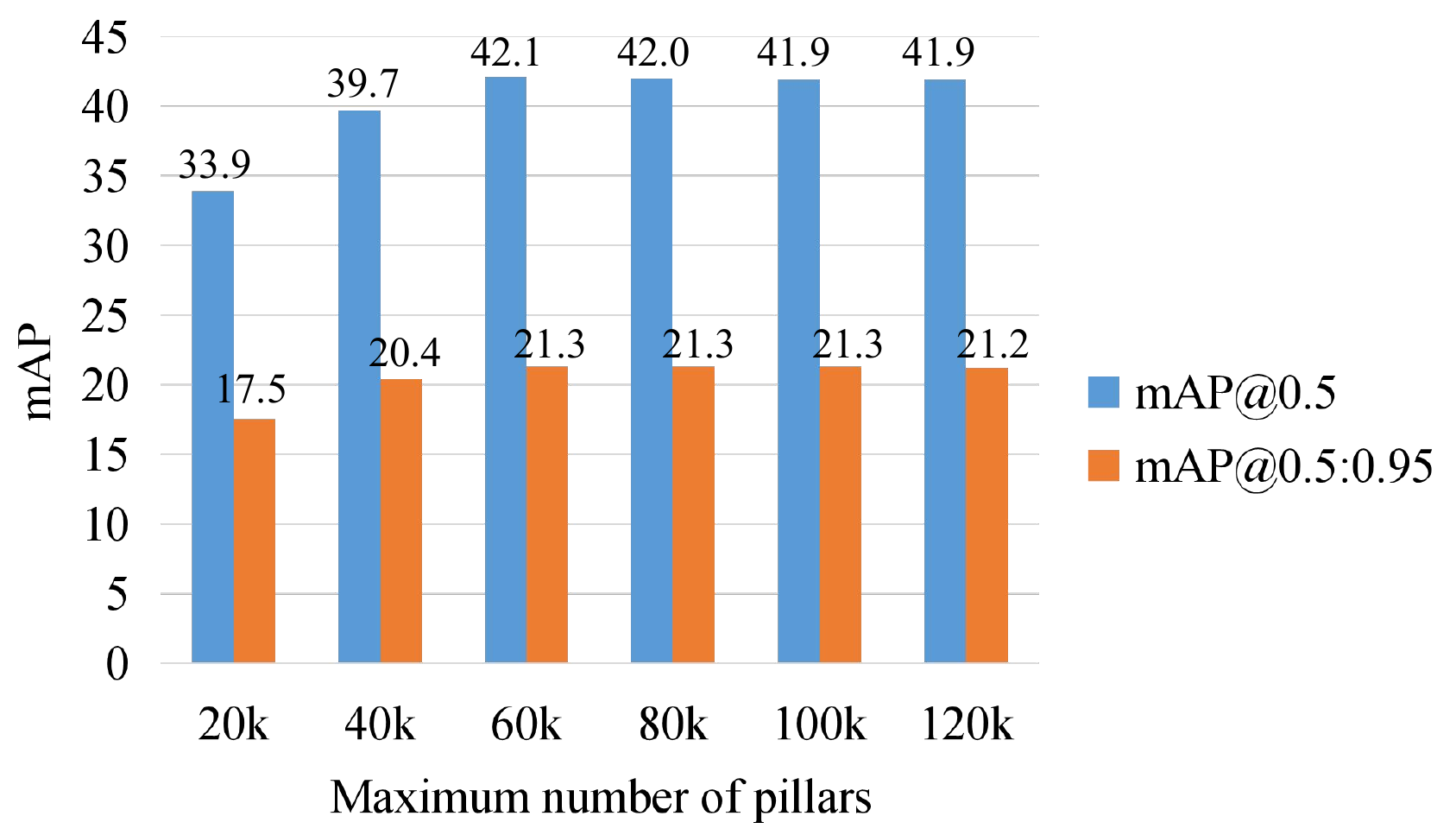}
    \centering
    \caption{Analysis on the maximum number of pillars.}
    \label{fig:abl_pillar}
\end{figure}

\noindent{\bfseries Analysis on the maximum number of pillars.} In EventPillars, a series of maximum number of pillars are used to aggregate information from event streams. To determine the appropriate number of pillars, we conduct a series experiments on different numbers of pillars from 20k to 120k. In this experiment, we train the ~\cite{perot2020learning} with EventPillars. We set the maximum number of pillars to 100k during training and set different number of pillars in inference phase. As shown in Fig.~\ref{fig:abl_pillar}, we achieve the best performance of 42.1\% mAP by setting maximum pillar number to 60k. It's obvious that if we decrease the pillar number from 40k to 20k, mAP will be dropped significantly. This is because there are still many events in 3D-space that are not covered by pillars. Similarly, when we keep increasing the number of pillars, all events will be utilized, leaving the extra pillars unused, thus mAP hardly increases anymore.

\section{Conclusion}
This work is targeted for event-based object detection. To make full use of information within events for the task of object detection, we propose a novel learnable event representation, namely EventPillars. It converts both positive and negative polarity of events into a set of pillars, generating a 3D tensor representation. Different from hand-crafted event representation methods, EventPillars can extract informative spatial-temporal features and can be trained together with object detector for high detection performance. In addition, an object detection framework designed for event streams called DMANet is presented. It leverages both long and short range memory modules to effectively aggregate spatial and temporal information for object detection. Extensive experiments demonstrate the effectiveness of the proposed methods and its superiority compared to other approaches.

\section{Acknowledgments}
The research was partially supported by the National  Natural Science Foundation of China, No. 62106036, 61725202, U1903215, 61829102, the Fundamental Research Funds for the Central University of China, DUT21RC(3)026,  Central Guidance on Local Science and Technology Development Fund of Liaoning Province no.2022JH6/100100026 and National Key R\&D Program of China under Grant No. 2018AAA0102001.

\bibliography{aaai23}

\end{document}